\documentclass[lettersize,journal]{IEEEtran}
\usepackage{amsmath,amsfonts}
\usepackage{algorithmic}
\usepackage{algorithm}
\usepackage{array}
\usepackage[caption=false,font=normalsize,labelfont=sf,textfont=sf]{subfig}
\usepackage{textcomp}
\usepackage{stfloats}
\usepackage{booktabs}
\usepackage{multirow}
\usepackage{url}
\usepackage{verbatim}
\usepackage{graphicx}
\usepackage{cite}
\usepackage{fontawesome}
\usepackage{orcidlink}
\usepackage{hyperref}
\usepackage{soul, color}

\begin{document}

\title{Unveiling the Unknown: Open Vocabulary Object Detection \\ with Scene Graphs}

\author{Yi Chen\textsuperscript{*}, Yinghao Lu\textsuperscript{*}, Zhehao Li\textsuperscript{*}, Chenchen Yan, Jiafei Wu \textsuperscript{\faEnvelope}, Chong Wang \textsuperscript{\faEnvelope},~\IEEEmembership{Member,~IEEE},\\ and Jiangbo Qian,~\IEEEmembership{Member,~IEEE}
\thanks{Manuscript received 17 March 2026; This work was supported by Zhejiang Provincial Natural
Science Foundation under Grant LMS26F030012; in part by Ningbo Municipal Natural Science Foundation of China under Grant 2022J114; in part by the National Natural Science Foundation of China under Grant 62271274; in part by Ningbo S\&T Project under Grant 2024Z004; and in part by Ningbo Major Research and Development Plan Project under Grant 2023Z225. }
\thanks{Yi Chen, Yinghao Lu, and Zhehao Li are with the Faculty of Electrical Engineering and Computer Science, Ningbo University, Ningbo, Zhejiang 315211, China (e-mail: 2411100286@nbu.edu.cn).}
\thanks{Chenchen Yan is with the Faculty of Computing, Georg-August-Universität Göttingen, Germany, majoring in Applied data science (e-mail: yancc0574@gmail.com).}
\thanks{Chong Wang and Jiangbo Qian are with the Merchants’ Guild Economics and Cultural Intelligent Computing Laboratory, Ningbo University, Ningbo, Zhejiang 315211, China, and also with the Faculty of Electrical Engineering and Computer Science, Ningbo University, Ningbo, Zhejiang 315211, China (e-mail: wangchong@nbu.edu.cn; qianjiangbo@nbu.edu.cn).}
\thanks{Jiafei Wu is with the School of Software Technology, Zhejiang University, Ningbo 315000, China (e-mail: {jcjiafeiwu@gmail.com}).}
\thanks{\textsuperscript{*}Yi Chen, Yinghao Lu and Zhehao Li contributed equally to this work.}
\thanks{\faEnvelope \ Corresponding Author: Jiafei Wu; Chong Wang.}
}

\markboth{Journal of \LaTeX\ Class Files,~Vol.~14, No.~8, August~2021}%
{Shell \MakeLowercase{\textit{et al.}}: A Sample Article Using IEEEtran.cls for IEEE Journals}

\IEEEpubid{0000--0000/00\$00.00~\copyright~2021 IEEE}

\maketitle

\begin{abstract}

Open-vocabulary object detection seeks to identify novel object categories that were not part of the training data. Many knowledge distillation-based approaches have shown promising performance by transferring knowledge from pre-trained vision-language models to object detection. However, these methods often overlook structured, image-specific relationships between objects, such as interactions and spatial arrangements. This oversight can significantly restrict the effectiveness of detecting novel categories. To address this issue, we propose a Scene-guided Relational Modeling detection framework. This framework utilizes scene graphs to capture structured semantic and spatial relationships between candidate regions and their contextual objects. It explicitly models interactions among neighboring regions and incorporates a Relation Attention Module to implicitly amplify the key relational cues extracted from the scene graph. Furthermore, we present a scene-based textual alignment branch that distills category knowledge from captions to guide relational alignment. This approach facilitates a seamless integration of visual relations with semantic information for enhanced detection performance. Comprehensive experiments show that our model achieves superior performance compared to other OVOD methods, improving the AP for novel categories on COCO and LVIS datasets.
\end{abstract}


\section{Introduction}
\label{sec:intro}
\IEEEPARstart{W}ith the widespread adoption of deep learning methods such as CNNs~\cite{ren2015faster} and Transformers~\cite{vaswani2017attention}, object detection has achieved remarkable progress. However, conventional detectors~\cite{ren2015faster,carion2020end} remain limited to a fixed set of categories and cannot recognize novel ones, restricting their applicability in diverse scenarios. To address this limitation, this paper focuses on the open-vocabulary object detection (OVOD) setting, where detectors are trained to identify objects from unseen categories.

\begin{figure}[t]
\begin{center}
\includegraphics[width=1.05\linewidth]{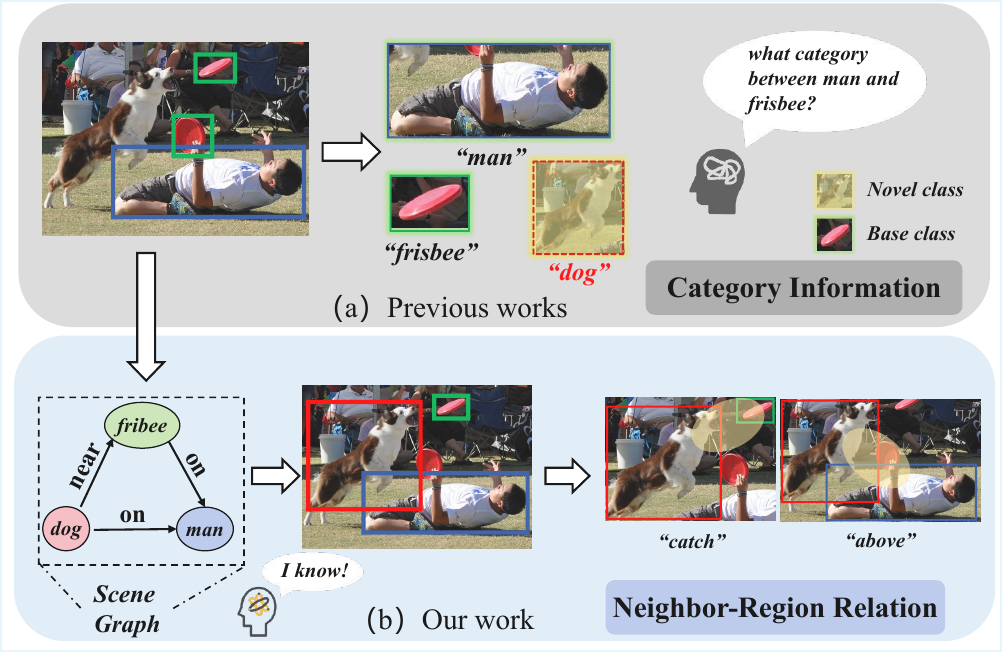}
\caption{(a) Previous works adopt various region partitioning strategies to capture intra-region category information for alignment with VLMs.  (b) In contrast, our proposed method enhances category associations by incorporating interactive relationships between neighboring regions.}
\label{intro}
\end{center}
\end{figure}


Currently, two main approaches are employed to tackle the OVOD problem.
The first utilizes external image-caption data as weak supervision to associate image regions with textual descriptions, thereby expanding the model’s semantic space~\cite{bravo2022localized,zhou2022detecting,lin2022learning,zhang2024exploring,kim2023contrastive}.
However, its reliance on simple region-text matching leads to coarse semantic alignment and limited scene understanding.
The second transfers knowledge from large-scale vision-language models (VLMs) via knowledge distillation~\cite{gu2021open,wu2023aligning,li2024learning}.
While these methods align regional features with VLMs to enhance category recognition (Fig.~\ref{intro}(a)), they often fail to fully exploit the compositional semantic structures inherent in VLMs.

\IEEEpubidadjcol

Further analysis reveals that the relationships between categories are crucial within the compositional structure of semantic concepts. Particularly in open-vocabulary scenarios, the detection of novel categories often relies on contextual links and semantic associations with known categories. However, existing methods overlook the interactive relationships between neighboring regions corresponding to the inputs of VLMs, as shown in Fig.~\ref{intro}(b). Intuitively, interactions between neighboring regions reflect category associations, and effectively modeling them is crucial for improving the generalization of object detectors.

To address the aforementioned shortcomings in modeling interactive relationships, an intuitive solution is to leverage external knowledge, such as the knowledge graph ConceptNet \cite{speer2017conceptnet}. However, such knowledge graphs are typically generic, and their retrieval results often contain a large amount of redundant information, making it difficult to provide precise and valuable semantic cues for OVOD. In contrast, we turn to a highly related task, scene graph generation (SGG) \cite{lu2016visual}, which generates a compact relational graph for each image to supplement visual cues with external knowledge. The relational information contained in scene graphs can be explicitly transmitted, helping the model to more accurately capture interactions between categories and more effectively generalize to novel categories.

Based on the above analysis, we propose a novel Scene-guided Relational Modeling (SRM) detection framework. 
It aims to improve novel category detection by modeling and enhancing the interactive relationships among categories. For each candidate region generated by the RPN, several neighboring proposals are randomly selected to form a neighbor-region, covering a wider range of visual concepts. A Neighbor-Region Relation Modeling (NR$^2$M) Branch is designed to explicitly learn relational information within the neighbor-region, combining with a Relation Attention Module (RAM) to enhance inter-region interactions.
This design captures both individual region features and cross-region relations, improving adaptability in open-vocabulary scenarios. Furthermore, a Scene-based Textual Alignment (STA) Branch based on image-caption data aligns relational features with textual semantics using a word retriever, facilitating fine-grained visual-textual alignment. To sum up, the contributions of our work are three-fold,
\begin{itemize}
\item 
A neighbor-region relation modeling (NR$^2$M) branch is proposed to explicitly model object interactions within neighboring regions, providing structured relational cues for open-vocabulary detection.
\item 
A relation attention module (RAM) is designed to implicitly enhances region-level interactions through a global-local attention mechanism, improving contextual reasoning and generalization.
\item 
A scene-based textual alignment (STA) branch is introduced to distill category knowledge from image caption pairs to guide alignment between visual regions and textual semantics.
\end{itemize}

\section{Related work}
\subsection{Open-Vocabulary Object Detection}
Object detection has long been a central task in computer vision, but traditional methods \cite{ren2015faster,redmon2016you,he2017mask,carion2020end,sun2021sparse} are limited by fixed object category sets, preventing them from adapting to unseen categories. To address this limitation, zero-shot object detection (ZSD) \cite{zero,zerohre,zerosynthesizing,zerotrans,zsddon,zsdgtnet} is introduced, aligning image region embeddings with category text embeddings through various strategies. Subsequently, open-vocabulary object detection (OVOD) further extends ZSD, achieving significant performance improvements under unified image-caption vocabulary knowledge. OVOD methods generally fall into two categories: region-aware training strategies and knowledge distillation methods. The former primarily introduces localization or contrastive losses to align regions with text based on image-caption pairs, as exemplified by VLDet \cite{lin2022learning}, which frames image-caption matching as a set matching problem between image region features and word embeddings.

The latter, knowledge distillation, directly distills (aligns) knowledge from vision-language models (VLMs) into detectors to enable open-vocabulary inference. 
For example, ViLD \cite{gu2021open} addresses the OVOD problem by leveraging CLIP’s image encoder for knowledge distillation. OADP \cite{wang2023object} enhances OVOD by introducing adaptive proposal cropping and extracting object-aware knowledge from VLMs, while BARON \cite{wu2023aligning} adopts a strategy of grouping contextually relevant regions and aligning them with the text encoder of VLMs to obtain holistic region representations.
Although these methods make notable progress, they do not fully explore the interactive relationships between neighboring regions corresponding to VLMs' inputs. In this paper, we propose a strategy that combines explicit modeling of relations within the neighbor-region and implicit enhancement of object interactions, thereby achieving effective generalization to novel categories.

\subsection{Vision-Language Models}
Vision-language models (VLMs), as large-scale pre-trained models, have gained significant attention at the intersection of computer vision and natural language processing. By learning from large-scale image-caption pairs, VLMs (\textit{e.g.}, CLIP \cite{radford2021learning} and ALIGN \cite{jia2021scaling}) align image and text representations and demonstrate impressive zero-shot image classification performance. These models, composed of an image encoder and a text encoder, are jointly trained to maximize similarity in the feature space. Early studies \cite{joulin2016,li2017} show that CNNs trained on caption prediction can learn representations comparable to those from ImageNet-21K \cite{deng2009imagenet}.

Recently, researchers have explored applying VLMs to dense prediction tasks like semantic segmentation and object detection \cite{xu2022groupvit,li2022language,gu2021open}. For example, MaskCLIP \cite{zhou2022extract} finds that CLIP’s image encoder, though not trained for detection, effectively captures object-level semantics and aligns pixel embeddings with text. This suggests VLMs can implicitly learn compositional visual semantics. However, the interactive relationships within such structures remain underexplored. To address this, we introduce a neighbor-region relation modeling branch to explicitly model interactions in neighbor regions and propose a relation attention module to further enhance them, thereby improving detection performance.

\subsection{Scene Graph Generation}
Early scene graph generation methods \cite{xu2017scene,lu2016visual} use pre-trained object detectors to extract bounding boxes for relation prediction without optimizing the detectors themselves. Recent works \cite{li2022integrating,li2021bipartite,li2022sgtr} jointly optimize object detection and relation prediction, forming the basis of our work. However, these approaches mainly focus on predicting relations from detection cues, with limited exploration of how relation cues can benefit detection. Some studies \cite{lyu2020vtgraphnet,liu2018structure} begin to leverage scene graph cues for improving object detection. For example, SIN \cite{liu2018structure} implicitly models object relationships without supervision for fixed-category detection, while SGMN \cite{yang2020graph} and vtGraphNet \cite{lyu2020vtgraphnet} convert complex 
referring expressions into scene graphs to facilitate reasoning.

In addition, several works \cite{he2022towards,zhong2021learning} explore open-vocabulary scene graph generation. Zhong 
\textit{et al}. \cite{zhong2021learning} uses weak supervision with image captions to learn open-vocabulary knowledge, and SVRP \cite{he2022towards} pre-trains on vision-language data for open-vocabulary relation recognition using prompt-based prediction. However, these methods focus on relation prediction and lack mechanisms for OVOD.
In this work, we propose a novel approach that utilizes scene graphs for open-vocabulary object detection. We introduce a neighbor-region relation modeling branch to explicitly model object interactions in neighboring regions, enhancing the detector’s generalization to novel categories.



\begin{figure*}[t]
\begin{center}
    \includegraphics[width=\linewidth]{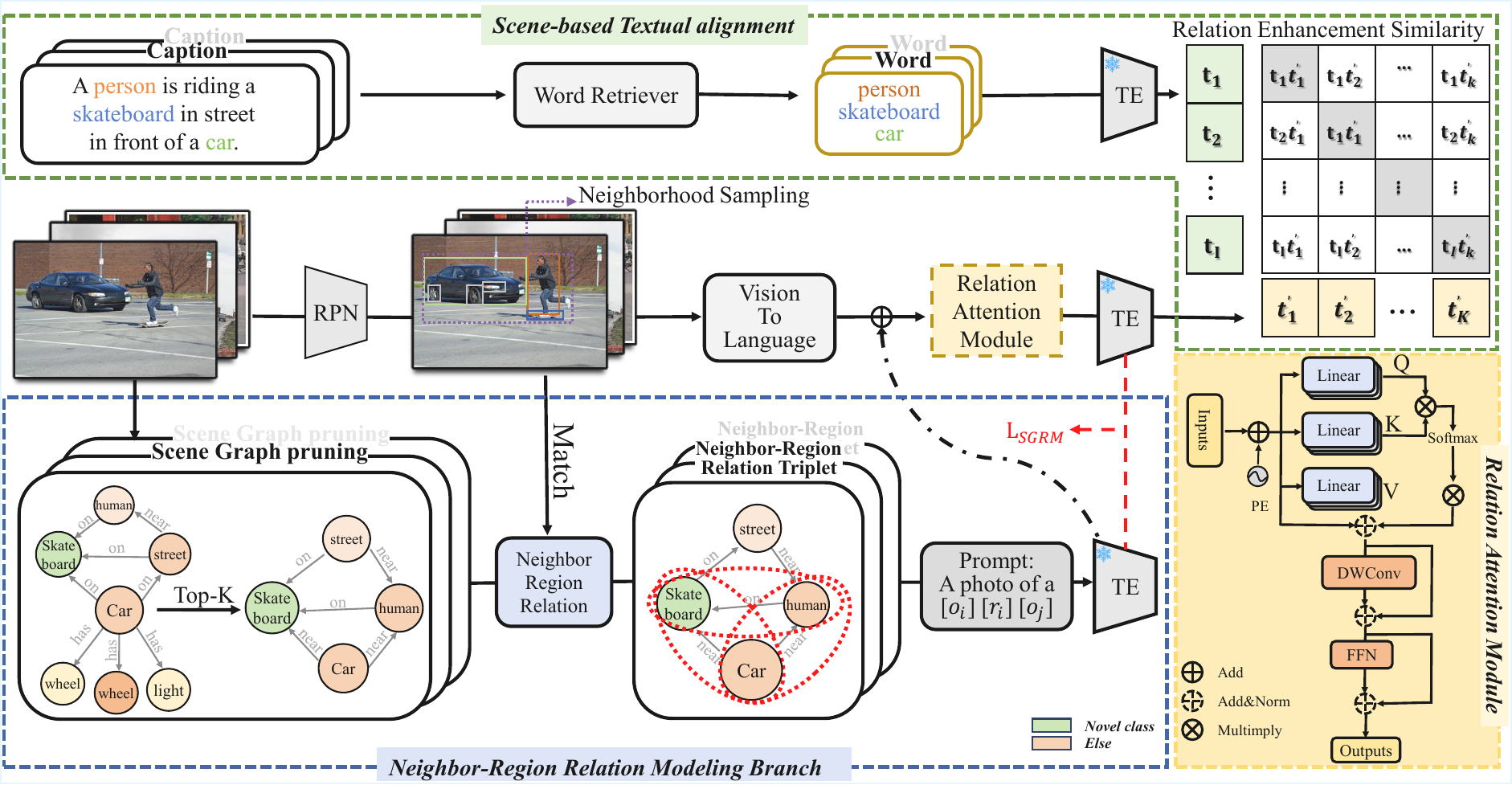}
 \caption{The overall architecture of the Scene-guided Relational Modeling (SRM) detection framework. It explicitly models and enhances the interactive relationships between categories within a neighbor-region to improve novel category detection. It integrates a neighbor-region relation modeling (NR$^2$M) branch for relational information and a relation attention module (RAM) to promote interactions among selected candidate boxes. Additionally, the scene-based textual alignment (STA) branch uses image-caption data to align relations with textual semantics. TE and RPN refer to the frozen text encoder and region proposal network, respectively. }
\label{main_fig1}
\end{center}
\end{figure*}

\section{Method}
\subsection{Preliminaries}\label{2.1}
For the open-vocabulary object detection (OVOD) task, we typically have an object detector trained on a detection dataset $D_{det}$, which contains the exhaustively annotated bounding-box labels for a set of base categories $C_{B}$. During inference, the categories in the testing set comprise both the base categories $C_{B}$ and novel categories $C_{N}$, \textit{i.e.}, $C_{test}=C_{B}\cup{C}_{N}$ and $C_{B}\cap{C}_{N}=\emptyset$.
Therefore, OVOD models need to address two key tasks: (1) accurately localizing all objects in the image, and (2) correctly classifying each object into a category in $C_{test}$.

As the same as many previous methods, we adopt the two-stage detection framework (\textit{e.g.} Faster R-CNN \cite{ren2015faster}) as the base detector.
In the first stage, a region proposal network (RPN) takes an image $I\in\mathbb{R}^{H\times W\times3}$ as the input and generates a set of class-agnostic candidate regions, 
represented by the bounding box $b_{i}=[x_{i},y_{i},w_{i},h_{i}]$, where $(x_i,y_i)$ denotes the coordinates of the top-left corner, $w_{i}$
and $h_{i}$ represents its width and height, respectively. In the second stage, each candidate region 
passes through a RoI-Align layer to obtain the region embedding, forming the set of region embeddings. 

Subsequently, following existing studies \cite{gu2021open,wu2023aligning}, the region embeddings are mapped into the word embedding space through a vision-to-language projection layer. The mapped word embeddings are then fed into the text encoder 
$\mathcal{T}(\cdot)$ to obtain the corresponding textual representations 
$\boldsymbol{e}_r$. To classify the region embeddings into categories from 
$C_{test}$, each category 
$c$ is appended to a prompt template (\textit{e.g.}, “A photo of a [CLS]”) and then encoded by the text encoder $\mathcal{T}(\cdot)$, resulting in the text embedding $\boldsymbol{t}_c$. Finally, the region embedding 
$\boldsymbol{r}$ is classified into category 
$c$ with the probability calculated as follows,
\begin{equation}p(\boldsymbol{r}, c)=\frac{\exp(cos(\boldsymbol{e}_r,\boldsymbol{t}_{c})/\tau)}{\sum_{i\in C_{test}}\exp(cos(\boldsymbol{e}_r,\boldsymbol{t}_{i})/\tau)},
\end{equation}
where $cos(\cdot,\cdot)$ denotes cosine similarity, $\tau$ is a temperature scaling factor.


\subsection{Scene-guided Relational Modeling}\label{2.2}
Most previous methods only implicitly capture the spatial relationships between neighboring regions corresponding to the inputs of VLMs, neglecting the more semantically meaningful interactions between categories within the scene. 
To exploit this missing information, we introduce the scene graph \cite{tang2020unbiased} as external knowledge to supplement visual cues. 
As shown in Fig.~\ref{main_fig1}, we first propose a neighbor-region relation modeling (NR$^2$M) branch to explicitly model the interactions among neighbor-region proposals. Building upon this, we propose a relation attention module (RAM) 
to further enhance interactions within the neighbor-region. Additionally, we introduce a scene-based textual alignment (STA) branch based on image-caption pairs, which employs a word retriever to extract category knowledge from captions to guide relational alignment, thereby improving the consistency between scene modeling and category recognition.


\subsubsection{Neighbor-Region Relation Modeling}
We first adopt a base model \cite{tang2020unbiased} 
pre-trained on the Visual Genome \cite{krishna2017visual} dataset to generate an initial scene graph 
${S}$ for each image, which can be formally defined as,
\begin{equation}
S=(V,E),\\
\end{equation}
where $V$ represents the set of detected object categories (with a total of 
$N$ objects) as,
\begin{equation}
V=\{o_i\}_{i=1}^N,
\end{equation}
and $E$ represents the set of relational triplets (with a total of 
$M$ relations) as,
\begin{equation}
E=\{e_{k}=<o_{i},r_{ij},o_{j}>\}_{k=1}^{M}.
\end{equation}
Here, $r_{ij}$ denotes the relation between object $o_{i}$ and $o_{j}$. For any two distinct object categories 
$o_{i}$ and $o_{j}$, there may exist two relations in opposite directions. Therefore, for 
$N$ object categories, there are 
$N(N-1)$ potential relations. 
It's important to note that if the scene is complex and contains numerous objects, the resulting scene graph will be enormous. This not only significantly increases computational costs but also introduces potential disruptions.
To address this issue, it becomes essential to prune the original graph before utilizing it. An intuitive approach is to select the 
$K$ most relevant potential relations and construct a sparse subgraph denoted as $S^{sub}$,
\begin{equation}
\begin{aligned}
&S^{sub}=TopK(E,p^{rel}),\\
&p^{rel}=Sigmoid(\boldsymbol{W_{r}}[f(o_{i}),f(r_{ij}),f(o_{j})]).
\end{aligned}
\end{equation}
Here, $o_i$ and $o_j$ denote the object nodes in the scene graph, and $r_{ij}$ denotes the semantic predicate between $o_i$ and $o_j$. The term $p^{rel}$ denotes the confidence score of the relational triplet, $\boldsymbol{W_{r}}$ is the learnable relationship matrix, and $f(\cdot)$ denotes the visual feature representation function for both objects and relational triplets.

Then, as shown in the middle panel of Fig.~\ref{main_fig1}, we construct relational triplets by matching the pruned subgraph \(S^{sub}\) to candidate boxes within a local neighborhood, treating \(S^{sub}\) as a query that ranks and selects the best matching candidates. Following \cite{wu2023aligning}, for each RPN proposal \(b\) we form a \(3\times3\) neighborhood \(P^{nbr}(b)=\{b\}\cup\{b+\delta_i\}_{i=1}^{8}\) (the eight offsets \(\delta_i\) are spatially adjacent, and out-of-bounds boxes are discarded). For each \(p\in P^{nbr}(b)\), we compute a subgraph box affinity \(s(p)=f(S^{sub},\phi(p))\) using the representation of \(S^{sub}\) and the RoI feature \(\phi(p)\). We keep the top-\(k\) boxes under \(s(\cdot)\) as matches. To enrich context and improve scale robustness, we sample \(G\) neighborhood instances at distinct scales and denote the \(g\)-th instance by \(P^{nbr}_g(b)\). We compute the class posterior matrix \(s_g^{nbr}\) for the \(g\)-th instance as,
\begin{equation}
s_g^{nbr}=\mathrm{Softmax}\!\Big(W_c\,[\,\phi(p)\,]_{p\in P^{nbr}_g(b)}\Big), 
\end{equation}
where \(W_c\) is the classification weight matrix, \(\phi(p)\) is the RoI feature of box \(p\) extracted from the backbone via RoIAlign, and the Softmax is applied column wise over classes. The predicted class for each box is the index of the largest entry in its column of \(s_g^{nbr}\). The category corresponding to the maximum probability can be viewed as the predicted label \(c_g^{nbr}\) for \(g\)-th instance candidate box.


Subsequently, we match the neighbor-region \(P_g^{nbr}\) with the filtered subgraph \(S_g^{sub}\) to ensure each candidate box is correctly associated with triplets in the subgraph. Specifically, we perform an IoU comparison between the bounding boxes \(b_g^{sub}\) in the subgraph and \(b_g^{nbr}\) in the neighbor-region. A pair is retained if the IoU exceeds the threshold $\theta$ and the category labels match,
\begin{equation}
\mathcal{E} = \{ E \in S_g^{sub} \mid \text{IoU}(b_g^{sub}, b_g^{nbr}) > \theta, c_g^{sub} = c_g^{nbr} \},
\end{equation}
where $b_g^{sub}$ denotes the bounding box in the pruned scene graph, $b_g^{nbr}$ denotes the bounding box in the neighbor-region, $c_g^{sub}$ is the category label of the scene-graph box, and $c_g^{nbr}$ is the predicted category label of the neighbor-region box. Using this strategy, we retrieve valid relational triplets \(\mathcal{E}\) from \(S_g^{sub}\) that match the candidate boxes. 

To encode the retrieved triplets \(\mathcal{E}\), we input each triplet into a prompt template (e.g., “A photo of a $[o_{i}][r_{ij}][o_{j}]$”) and feed the generated text into the text encoder \(\mathcal{T}(\cdot)\), obtaining the corresponding relational embeddings \(\boldsymbol{R}^{nbr}\).

\subsubsection{Relation Attention Module}

As external knowledge, the introduction of scene graphs provides semantic associations for object categories. It can be further enhanced by fusing with visual clues.
Specifically, for the previously generated neighbor-region ${P_g^{nbr}}$, we follow the open-vocabulary detection process by inputting it into an RoI-Align layer to extract the region features. These features are then mapped to the word embedding space through a vision-to-language (V2L) projection layer to obtain the neighborhood embeddings. Subsequently, we concatenate the relational embeddings with the neighbor-region embeddings to obtain the fused neighbor-region relation embeddings $\widetilde{\boldsymbol{R}}^{nbr}$, as shown in the middle part of Fig.~\ref{main_fig1}.

Then, a relation attention module (RAM) is used to strengthen key interaction cues by introducing a global-local attention mechanism, as presented in the lower right corner of Fig.~\ref{main_fig1}.
The multi-head self-attention (MSA) is performed to obtain the global attention of various neighbor-regions. 
By injecting positional encoding (PE) in $\widetilde{\boldsymbol{R}}^{nbr}$, the query $\boldsymbol{Q}$, key $\boldsymbol{K}$, and value $\boldsymbol{V}$ are obtained as,
\begin{equation}
\boldsymbol{Q}=\widetilde{\boldsymbol{R}}^{nbr}\cdot \boldsymbol{W_{Q}^{\top}},\boldsymbol{K}=\widetilde{\boldsymbol{R}}^{nbr}\cdot \boldsymbol{W_{K}^{\top}},\boldsymbol{V}=\widetilde{\boldsymbol{R}}^{nbr}\cdot \boldsymbol{W_{V}^{\top}},
\end{equation}
where $\boldsymbol{W_{Q}^{\top}}$, $\boldsymbol{W_{K}^{\top}}$, and $\boldsymbol{W_{V}^{\top}}$ represent the corresponding weight matrices. 
The final attention can be formulated as,
\begin{equation}
Attention(\boldsymbol{Q},\boldsymbol{K},\boldsymbol{V})=softmax\left(\frac{\boldsymbol{Q}\boldsymbol{K}^\top}{\sqrt{d_{word}}}\right)\boldsymbol{V},
\end{equation}
where $d_{word}$ is the dimension of the word embeddings. 

On the other hand, 
the fine-grained relational interactions within each neighbor-region is extracted using depthwise separable convolutions (DWC) as the local attention.
A feed-forward network (FFN) with GELU \cite{hendrycks2016gaussian} and linear layers is followed, while residual connections and layer normalization are incorporated. Through the above process, RAM combines global and local attention mechanisms to construct a network component with strong interaction capability, outputting the neighbor-region relation embeddings $\hat{\boldsymbol{R}}^{nbr}$ with rich relational semantics.

\subsubsection{Scene-based Textual Alignment}

The objective of OVOD is to identify novel categories, which necessitates that the extracted neighbor-region relation embeddings $\hat{\boldsymbol{R}}^{nbr}$ be constrained by object categories rather than merely their interrelations. To achieve this, a scene-based textual alignment (STA) branch is conducted to amplify category-specific information over relational semantics. This strategy guides the relational modeling process to more accurately focus on semantically interacting objects.
Specifically, to leverage fine-grained knowledge embedded in the caption text, a word retriever \cite{schuster2015generating} is employed to extract category words $W=\{W_{1},\ldots,W_{i},\ldots,W_{|W|}\}$ from the caption text, where $|W|$ represents the number of words in the caption. As illustrated in the upper part of Fig.~\ref{main_fig1}, these category words correspond to the object categories mentioned in the caption text. 
Each word is embedded through the text encoder $\mathcal{T}(\cdot)$, generating word embeddings $\boldsymbol{t}^{word}$ as,
\begin{equation}
\boldsymbol{t}^{word}=\mathcal{T}(W).
\end{equation}

Meanwhile, the neighbor-region relation embeddings 
$\hat{\boldsymbol{R}}^{nbr}$ are also fed into the text encoder $\mathcal{T}(\cdot)$ to obtain a text embedding ${\boldsymbol{t}}^{nbr}$ that contains interaction relationships within the neighbor-region. Finally, we apply a contrastive learning 
to align ${\boldsymbol{t}}^{word}$ with ${\boldsymbol{t}}^{nbr}$ 
as, with positive pairs constructed between the neighbor-region relation embedding and its corresponding category word retrieved from the image caption. This process is formalized as, 
\begin{equation}
L_{SRM}=-\frac{1}{2}\sum_{k=0}^{n_G-1}(\log(p_k^{n,w})+\log(p_k^{w,n})),
\end{equation}
where $p_k^{n,w}$ and $p_k^{w,n}$ are computed as follows,
\begin{equation}
\begin{gathered}
p_{k}^{n,w}=\frac{\exp(\langle\boldsymbol{t}_k^{nbr},\boldsymbol{t}_k^{word}\rangle/\tau)}{\sum_{l=0}^{n_G-1}\exp(\langle\boldsymbol{t}_k^{nbr},\boldsymbol{t}_l^{word}\rangle/\tau)},\\p_{k}^{w,n}=\frac{\exp(\langle\boldsymbol{t}_k^{word},\boldsymbol{t}_k^{nbr}\rangle/\tau)}{\sum_{l=0}^{n_G-1}\exp(\langle\boldsymbol{t}_k^{word},\boldsymbol{t}_l^{nbr}\rangle/\tau)}.
\end{gathered}
\end{equation}
Here,${\boldsymbol{t}}^{nbr}$ represents the text embedding encoding interactions among neighboring regions, while ${\boldsymbol{t}}^{word}$ represents the word embeddings. The loss encourages positive pairs to be close and negative pairs to be distant.

During training, we jointly optimize this contrastive loss with conventional object detection losses. Therefore, the total loss function of the proposed method in this work is defined as,
\begin{equation}
L_{total}=L_{SRM}+L_{cls}+L_{reg}+L_{rpn},
\end{equation}
where $L_{cls}$, $L_{reg}$, and $L_{rpn}$ correspond to the classification loss, bounding box regression loss, and region proposal loss, respectively.

\begin{table}[t]
\centering
\caption{Comparison results on OV-COCO benchmark. \textbf{Bold} and \underline{underline} items indicate the best and second-best results, respectively. 
}
\label{ov-coco}
\resizebox{\columnwidth}{!}{
\begin{tabular}{lcccc}
\toprule
Method & Backbone  & $\mathrm{AP}_{50}^{\mathrm{novel}}$ & $\mathrm{AP}_{50}^{\mathrm{base}}$ & $\mathrm{AP}_{50}^{\mathrm{all}}$ \\
\midrule
\multicolumn{5}{l}{\textit{Def-DETR based methods}} \\
\midrule
OV-DETR~\cite{zang2022open} & RN50-C4  & 29.4 & \underline{61.0} & 52.7 \\
Prompt-OVD~\cite{feng2022promptdet} & ViT-B/16 & \underline{20.6} & \textbf{63.5} & \underline{54.9} \\
DK-DETR~\cite{li2023distilling} & RN50 & \textbf{32.3} & - & \textbf{61.1} \\
\midrule
\multicolumn{5}{l}{\textit{Faster R-CNN based methods}} \\
\midrule
OVR-CNN~\cite{zareian2021open} & RN50  & 22.8 & 46.0 & 39.9 \\
ViLD~\cite{gu2021open} & RN50-FPN &  27.6 & \textbf{59.5} & \textbf{51.2 }\\
Detic~\cite{zhou2022detecting}  & RN50-C4 & 27.8 & 47.1 & 45.0 \\
OADP~\cite{wang2023object} & RN50 & 30.0 & 53.3 & 47.2 \\
ProxyDet~\cite{jeong2024proxydet} & RN50-C4 & 30.4 & 52.6 & 46.8 \\
VLDet~\cite{lin2022learning} & RN50-C4  & 32.0 & 50.6 & 45.8 \\
BARON~\cite{wu2023aligning} & RN50-C4 & 33.1 & 54.8 & 49.1 \\
RALF~\cite{kim2024retrieval} & RN50-C4 & 33.4 & 54.5 & 49.0 \\
CODet\cite{li2025codet} & RN50-C4 & \underline{33.9} & 52.3 & 47.6 \\
\textbf{Ours} & RN50-C4 &  \textbf{36.9} & \underline{54.9} & \underline{50.2} \\
\bottomrule
\end{tabular}
}
\end{table}

\begin{table}[t]
\centering
\caption{Comparison results on OV-LVIS benchmark.}
\label{ov-lvis}
\begin{tabular}{lccccc}
\toprule
Method & Backbone & $\mathrm{AP}_{\mathrm{r}}$ & $\mathrm{AP}_{\mathrm{c}}$ & $\mathrm{AP}_{\mathrm{f}}$ & $\mathrm{AP}_{\mathrm{all}}$ \\
\midrule
ViLD~\cite{gu2021open} & RN50-FPN & 16.3 & 21.2 & 31.6 & 24.4 \\
RegionCLIP~\cite{zhong2022regionclip} & RN50-C4 & 17.1 & 27.4 & 34.0 & 28.2 \\
ProxyDet~\cite{jeong2024proxydet} & RN50-C4 & 18.9 & - & - & 30.1 \\
Detic~\cite{zhou2022detecting} & RN50-C4 & 19.5 & - & - & \underline{30.9} \\
DetPro~\cite{du2022learning} & RN50-FPN & 19.8 & 25.6 & 28.9 & 29.6 \\
RALF~\cite{kim2024retrieval} & RN50-C4 & 21.1 & 25.7 & 29.2 & 26.3 \\
MMC-Det~\cite{xu2024exploring} & RN50-FPN & 21.1 & \textbf{30.9} & \textbf{35.5} & \textbf{31.0} \\
OADP~\cite{wang2023object} & RN50 & 21.9 & 28.4 & 32.0 & 28.7 \\
LBP~\cite{li2024learning} & RN50-FPN & 22.2 & 28.8 & 32.4 & 29.1 \\
BARON~\cite{wu2023aligning} & RN50-FPN & \underline{23.2} & 29.3 & 32.5 & 29.5 \\

\textbf{Ours} & RN50-FPN & \textbf{24.1} & \underline{30.2} & \underline{34.9} & 30.8 \\
\bottomrule
\end{tabular}
\end{table}

\begin{table}[t]
\centering
\caption{Comparison results on the transfer dataset. The OV-LVIS trained model is evaluated on COCO and Objects365.}
\label{transfer}
\setlength{\tabcolsep}{4pt}

\begin{tabular}{lcccccc}
\toprule
\multirow{2}{*}{Method} & \multicolumn{3}{c}{COCO} & \multicolumn{3}{c}{Objects365} \\
\cmidrule(lr){2-4} \cmidrule(lr){5-7}
 & AP$_{\text{all}}$ & AP$_{50}$ & AP$_{75}$ & AP$_{\text{all}}$ & AP$_{50}$ & AP$_{75}$ \\
\midrule
Supervised~\cite{gu2021open} & 46.5 & 67.6 & 50.9 & 25.6 & 38.6 & 28.0 \\
\midrule
F-VLM~\cite{kuo2022} & 32.5 & 53.4 & 34.6 & 11.9 & 19.2 & 12.6 \\
DetPro~\cite{du2022learning} & 34.9 & 53.8 & 37.4 & 12.1 & 18.8 & 12.9 \\
BARON~\cite{wu2023aligning} & 36.2 & \underline{55.7} & 39.1 & \underline{13.6} & \underline{21.0} & \underline{14.5} \\
ViLD~\cite{gu2021open} & \underline{36.6} & 55.6 & \underline{39.8} & 11.8 & 18.2 & 12.6 \\
\textbf{Ours} & \textbf{37.5} & \textbf{56.4} & \textbf{40.7} & \textbf{13.9} & \textbf{21.6} & \textbf{14.9} \\
\bottomrule
\end{tabular}
\end{table}

\section{Experiments}
\subsection{Experimental Setup}
\textbf{Datasets.}
We conduct experiments on two popular OVOD benchmark datasets based on COCO \cite{lin2014microsoft} and LVIS \cite{gupta2019lvis}. For the COCO dataset, we follow the open-vocabulary setting defined by \cite{zareian2021open}, where the categories are divided into 48 base categories and 17 novel categories.
We train the detector using the ground-truth annotations of the base category and evaluate it on the validation set. This configuration is referred to as OV-COCO. Furthermore, for image-caption data, following existing works \cite{lin2022learning,zareian2021open,zhou2022detecting}, we utilize the COCO Caption \cite{chen2015microsoft} training set for experiments on OV-COCO, which provides 5 human-generated captions for each image.

Regarding the LVIS dataset, it is essentially a long-tailed dataset, with categories divided into three distinct levels: frequent, common, and rare. Following standard practice \cite{gu2021open,zhong2022regionclip}, we set the 866 common and frequent categories in LVIS as base categories and the 337 rare categories as novel categories. We remove the novel class annotations during training and predict all categories during testing. This configuration is referred to as OV-LVIS. Furthermore, we choose CC3M \cite{sharma2018conceptual} as the source of image-caption pairs, which contains 2.8 million free-form image-caption pairs crawled from the Web.

\textbf{Evaluation metrics.} 
For completeness, we evaluate the detection performance on both base and novel categories. Following the prior work \cite{zareian2021open,wu2023aligning}, for OV-COCO, we report the box AP at an IoU threshold of 0.5. $\mathrm{AP}_{50}^{\mathrm{novel}}$, $\mathrm{AP}_{50}^{\mathrm{base}}$ and $\mathrm{AP}_{50}^{\mathrm{all}}$ are the metrics for the novel, base, and all (novel + base) categories, respectively. For OV-LVIS, we report the mask AP averaged over IoUs from 0.5 to 0.95. $\mathrm{AP}_{\mathrm{r}}$, $\mathrm{AP}_{\mathrm{c}}$, $\mathrm{AP}_{\mathrm{f}}$ and $\mathrm{AP}_{\mathrm{all}}$ are the metrics for rare, common, frequent, and all categories, respectively. $\mathrm{AP}_{50}^{\mathrm{novel}}$ and $\mathrm{AP}_{\mathrm{r}}$ are the primary metrics for evaluating the open-vocabulary object detection performance on OV-COCO and OV-LVIS, respectively.

\textbf{Implementation details.} 
Our model is built upon the Faster R-CNN~\cite{ren2015faster} framework using PyTorch~\cite{paszke2019pytorch}. The text encoder adopts the ViT-B/32 CLIP~\cite{dosovitskiy2020image} model with frozen weights. For scene graph generation, we apply different strategies for COCO and LVIS. On COCO, we use a Visual Genome~\cite{krishna2017visual} pre-trained scene graph model to generate relational graphs. Moreover, we set \(\theta = 0.5\) for the IoU threshold in the matching process.
 For LVIS, due to the category discrepancy between its long-tailed distribution and Visual Genome, we align categories via the CLIP text encoder, allowing soft matching for pairs with similarity above 0.6 to improve relational coverage and semantic consistency. For OV-LVIS, we use a ResNet50-FPN~\cite{lin2017feature} backbone initialized with SoCo~\cite{wei2021aligning} pre-trained weights and train with SGD (lr=0.01, momentum=0.9, weight decay=$2.5\times10^{-5}$). For OV-COCO, we employ a ResNet50-C4 backbone with SGD (lr=$5\times10^{-3}$, momentum=0.9, weight decay=$10^{-4}$). All experiments are trained for 180k iterations on 4 NVIDIA 4090 GPUs, with linear warm-up for 5k iterations and learning rate decay by a factor of 10 at 120k and 180k iterations.

\begin{table*}[t]
\centering
\begin{minipage}{0.48\linewidth}
\centering
\caption{Ablation study on components of the NRE model on OV-COCO benchmark.}
\label{ablation1}
\small
\begin{tabular}{llllll}
\toprule
NR$^2$M & RAM & STA & $\mathrm{AP}_{50}^{\mathrm{novel}}$ & $\mathrm{AP}_{50}^{\mathrm{base}}$ & $\mathrm{AP}_{50}^{\mathrm{all}}$ \\
\midrule
- & - & - & 32.0 & 53.4 & 47.8 \\
\checkmark & - & - & 31.7 & 53.8 & 48.1 \\
- & \checkmark & - & 32.8 & 54.1 & 48.5 \\
- & - & \checkmark & 33.5 & 53.8 & 48.4 \\
\checkmark & \checkmark & - & 33.2 & 54.3 & 48.7 \\
\checkmark & - & \checkmark & 34.6 & 54.0 & 48.9 \\
- & \checkmark & \checkmark & 35.7 & 54.7 & 49.7 \\
\checkmark & \checkmark & \checkmark & \textbf{36.9} & \textbf{54.9} & \textbf{50.2} \\
\bottomrule
\end{tabular}
\end{minipage}
\hfill
\begin{minipage}{0.48\linewidth}
\centering
\caption{Ablation study on RAM components on OV-COCO bechmark.}
\label{ablation2}
\small
\begin{tabular}{lllllll}
\toprule
SA & DWConv & FFN & PE & $\mathrm{AP}_{50}^{\mathrm{novel}}$ & $\mathrm{AP}_{50}^{\mathrm{base}}$ & $\mathrm{AP}_{50}^{\mathrm{all}}$ \\
\midrule
- & - & - & - & 34.6 & 54.0 & 48.9 \\
\checkmark & - & - & - & 35.7 & 54.2 & 49.3 \\
\checkmark & \checkmark & - & - & 36.3 & 54.6 & 49.8 \\
\checkmark & \checkmark & \checkmark & - & 36.5 & 54.8 & 50.0 \\
\checkmark & \checkmark & \checkmark & \checkmark & \textbf{36.9} & \textbf{54.9} & \textbf{50.2} \\
\bottomrule
\end{tabular}
\end{minipage}
\end{table*}

\subsection{Comparisons with Other Methods}
To ensure a fair comparison, we divide the methods into two groups based on the detection framework they employ: one group uses the Faster R-CNN framework, while the other uses the DETR framework. It can be observed that, under the same backbone network and detection framework (Faster R-CNN), the proposed SRM model achieves a performance of 36.9\% in $\mathrm{AP}_{50}^{\mathrm{novel}}$, improving by 3.5\% over the current second-best method, RALF \cite{kim2024retrieval}. Compared to the baseline model (BARON \cite{wu2023aligning}), our model improves by 3.8\%, 0.1\%, and 1.1\% in $\mathrm{AP}_{50}^{\mathrm{novel}}$, $\mathrm{AP}_{50}^{\mathrm{base}}$, and $\mathrm{AP}_{50}^{\mathrm{all}}$, respectively.
Furthermore, compared to methods based on the DETR framework, such as Prompt-OVD \cite{feng2022promptdet} and DK-DETR \cite{li2023distilling}, the proposed model consistently outperforms in terms of novel category detection but shows slightly weaker performance on base categories. Analyzing the reason, models based on the DETR framework typically employ the Transformer architecture, which requires a large number of training parameters and may result in a bias towards base category predictions during training. CODet \cite{li2025codet} further improved open-vocabulary detection by leveraging visual–textual co-occurrence knowledge, offering a complementary direction to architecture-based approaches. However, in open-vocabulary object detection, the ability to recognize novel categories is crucial, further validating the effectiveness of the object relation modeled using scene graphs.

As shown in Table \ref{ov-lvis}, our method is also evaluated on the OV-LVIS dataset. To reduce training time and computational resource consumption, the instance segmentation task is excluded from this evaluation. In the object detection task, the proposed SRM model consistently outperforms existing methods in detecting novel categories. 
Compared to the baseline method \cite{wu2023aligning}, the SRM model achieves a 0.9\% improvement in the 
$\mathrm{AP}_{\mathrm{r}}$ metric. Additionally, in base category detection (including common and frequent categories), the proposed model demonstrates strong competitiveness, achieving 30.2\%, 34.9\%, and 30.8\%, respectively. These experimental results not only validate the open-vocabulary reasoning capability of the SRM model for novel categories, but also indicate that, on the OV-LVIS benchmark, the proposed SRM model improves novel category detection performance without significantly compromising performance on base categories.

\subsection{Transfer to Other Datasets}
To further verify the open-vocabulary generalization capability of the proposed model, we evaluate its transfer performance on other object detection datasets, as shown in Table.~\ref{transfer}. The results indicate that our method exhibits significantly better transfer performance on the COCO \cite{lin2014microsoft} dataset compared to other baseline approaches, ranking only behind the upper-bound performance achieved through fully supervised training (Row 1).
When transferring to the Objects365 \cite{shao2019objects365} dataset, the NRE model achieves performance improvements of 0.3\%, 0.6\%, and 0.4\% on the key metrics of $\mathrm{AP}_{\mathrm{all}}$, $\mathrm{AP}_{\mathrm{50}}$, and $\mathrm{AP}_{\mathrm{75}}$, respectively, compared to the current state-of-the-art method BARON \cite{wu2023aligning}. These results demonstrate that even for object categories originating from different data distributions, the NRE model is still able to maintain strong detection performance and generalization capability.

\begin{table*}[t]
\centering
\begin{minipage}{0.48\linewidth}
\centering
\caption{Ablation study on vocabulary size on OV-COCO benchmark.}
\label{ablation4}
\small
\begin{tabular}{llll}
\toprule
Vocabulary & $\mathrm{AP}_{50}^{\mathrm{novel}}$ & $\mathrm{AP}_{50}^{\mathrm{base}}$ & $\mathrm{AP}_{50}^{\mathrm{all}}$ \\
\midrule
COCO Caption & 33.2 & 54.3 & 48.7 \\
Category words from Caption & \textbf{36.9} & \textbf{54.9} & \textbf{50.2} \\
\bottomrule
\end{tabular}
\end{minipage}
\hfill
\begin{minipage}{0.48\linewidth}
\centering
\caption{Ablation study on multi-layer stacked RAM on OV-COCO benchmark.}
\label{ablation3}
\small
\begin{tabular}{cccc}
\toprule
Layers & $\mathrm{AP}_{50}^{\mathrm{novel}}$ & $\mathrm{AP}_{50}^{\mathrm{base}}$ & $\mathrm{AP}_{50}^{\mathrm{all}}$ \\
\midrule
1 & 35.7 & \textbf{55.1} & 50.0 \\
3 & \textbf{36.9} & 54.9 & \textbf{50.2} \\
5 & 36.2 & 54.5 & 49.7 \\
\bottomrule
\end{tabular}
\end{minipage}
\end{table*}

\begin{figure*}[t]
  \centering
  \begin{minipage}[t]{0.48\textwidth}
    \centering
    \includegraphics[width=\linewidth]{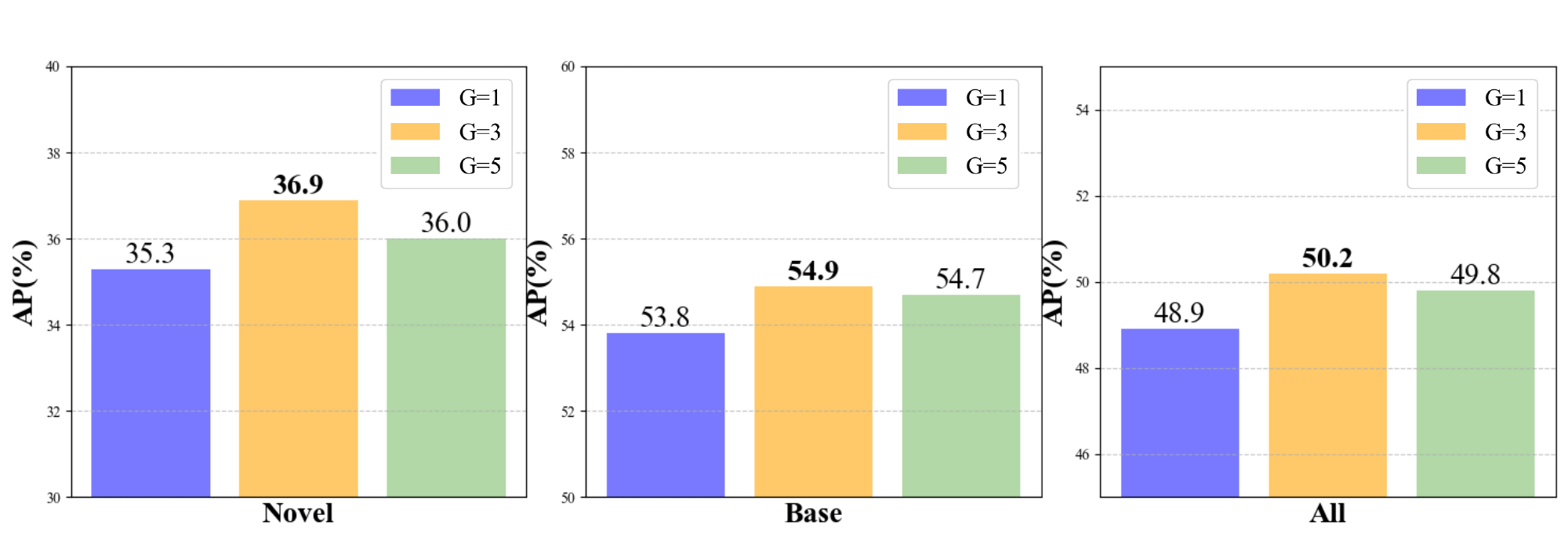}
    \caption{Ablation study on the number of sampled groups (Top-$G$) on OV-COCO benchmark.}
    \label{visual3}
    
  \end{minipage}
  \hfill
  \begin{minipage}[t]{0.48\textwidth}
    \centering
    \includegraphics[width=\linewidth]{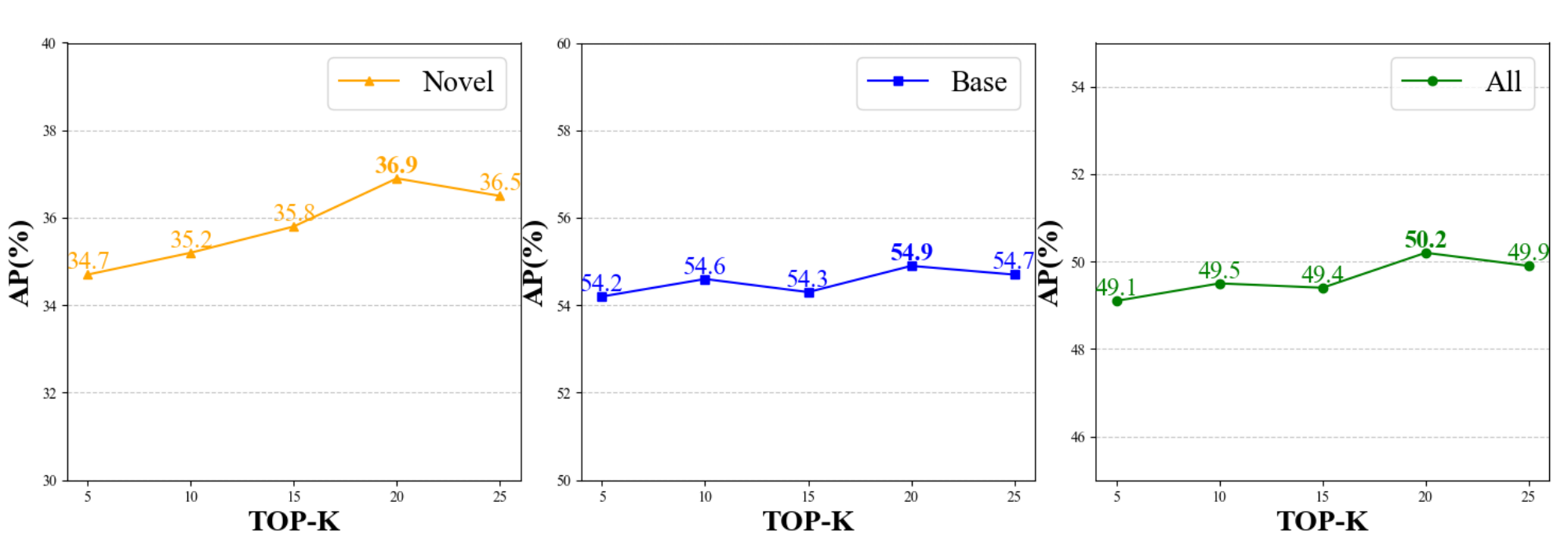}
    \caption{Ablation study on most relevant potential relations (Top-$K$) on OV-COCO benchmark.}
    \label{visual4}
  \end{minipage}
\end{figure*}


\begin{figure*}[t]
  \centering
  \includegraphics[width=\linewidth]{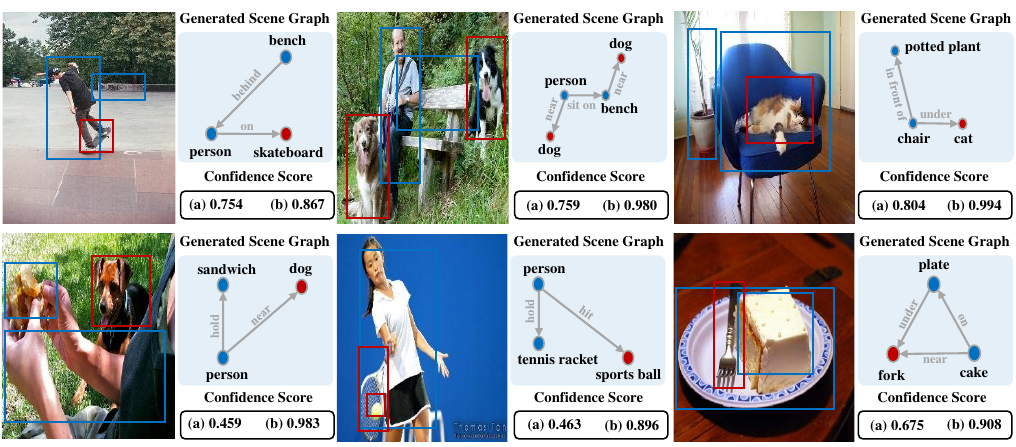}
  \caption{Qualitative results to show the effectiveness of the NR$^2$M. The red bounding boxes (nodes) represent novel categories,
  while the blue bounding boxes (nodes) represent base categories. \textbf{(a):} w/o NR$^2$M. \textbf{(b):} w/ NR$^2$M.}
  \label{visual5}
\end{figure*}


\begin{figure*}[t]
  \centering
  \includegraphics[width=\linewidth]{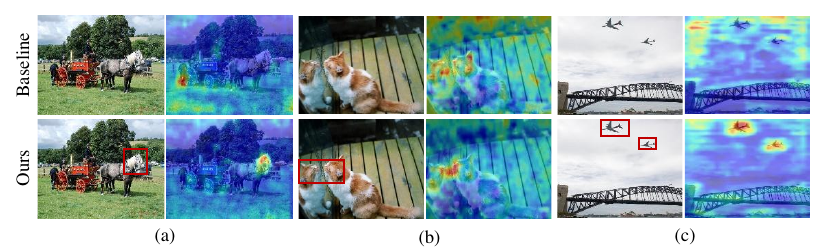}
  \caption{Qualitative comparison results on OV-COCO. For each group, red boxes indicate the novel categories and the feature map responses to queried novel categories.
  From (a) to (c), the queried novel categories are “horse”, “cat”, and “airplane”.}
  \label{visual6}
\end{figure*}

\subsection{Ablation Studies}
More ablation studies are conducted to validate the effectiveness of the proposed SRM model on the OV-COCO benchmark.

\noindent{\textbf{Neighbor-Region Relation Modeling.}} As shown in Table~\ref{ablation1}, 
when using NR$^2$M alone, the performance on base categories improves significantly, while the performance on novel categories declines. This phenomenon indicates that relying solely on relation knowledge generated by NR$^2$M may introduce noise, which adversely affects the detection of novel categories. Further analysis reveals that using RAM and STA independently results in performance improvements of 0.8\% and 1.5\% on novel categories, respectively, demonstrating the effectiveness of these two modules.
Notably, combining STA with NR$^2$M and RAM improves performance on novel categories by 1.1\% and 2.2\%, respectively, compared to STA alone, indicating a synergistic effect between STA and the other modules. When all three modules NR$^2$M, RAM, and STA are applied together, performance increases by 1.2\%, providing strong evidence of their complementarity.


\noindent{\textbf{Relation Attention Module.}}
To assess the impact of each component in the proposed Relation Attention Module (RAM), we report the results in Table~\ref{ablation2}. As described in the methodology, RAM consists of four key components: self-attention (SA), depthwise separable convolution (DWC), feed-forward network (FFN), and positional encoding (PE).
The results show that incorporating SA into the model significantly improves the performance on novel categories, with $\mathrm{AP}_{50}^{\mathrm{novel}}$ increasing from 34.6\% to 35.7\%, demonstrating its effectiveness in enhancing novel category detection. Subsequently, with the addition of DWC, the $\mathrm{AP}_{50}^{\mathrm{novel}}$ metric is further improved by 0.6\%. Finally, by integrating the FFN and PE modules, the performance on novel categories reaches 36.9\%.
In terms of base category detection, the model also exhibits noticeable improvements, with the $\mathrm{AP}_{50}^{\mathrm{base}}$ metric increasing from 54.6\% to 54.9\%.

\noindent{\textbf{The number of sampled groups (Top-$G$).}}
We conduct another ablation study on the number of neighborhood sampling groups (Top-$G$) to evaluate the impact of this parameter on detection performance. The results are presented in Fig.~\ref{visual3}.
It can be observed that as the number of sampling groups increases from 1 to 3, the performance on novel, base, and overall categories reaches its peak. However, when the number increases to 5, performance on novel categories begins to decline.
These findings suggest that sampling three neighborhood groups per candidate region allows the detector to learn comprehensive representations, thereby enhancing its generalization to novel categories.

\noindent{\textbf{The most relevant potential relations (Top-$K$).}}
From the Fig.~\ref{visual4}, it can be observed that as the value of $K$ increases from 5 to 20, the performance on novel categories consistently improves. At the same time, the performance on base and overall categories also gradually increases, with the best overall performance achieved when K is set to 20. However, further increasing $K$ leads to a decline in performance.
These results indicate that $K$ = 20 is the optimal setting for scene graph sparsification, as it provides high-quality structured priors for the subsequent relation attention and fine-grained alignment processes.



Novel categories are often correctly identified through their relationships with base categories. For example, in the first row, “skateboard” is linked to “person” via the “on” relation. After introducing the NR$^2$M branch, the model’s confidence in detecting novel categories improves significantly, demonstrating that NR$^2$M provides structured contextual information, enabling the model to perceive and localize novel categories through interactions with base categories, even without explicit supervision.

\subsection{Qualitative Analysis}
To qualitatively analyze detection performance, we compare our model with the baseline model \cite{wu2023aligning} under the OV-COCO setting, as illustrated in Fig.~\ref{visual6}. To visualize the feature map responses, we adopt Grad-CAM++ \cite{chattopadhay2018grad} to generate heatmaps corresponding to the regions activated by the novel categories.
Notably, the model trained with NRE generates clear and focused activation regions for novel categories. For example, in (a), the first column shows a highly concentrated heatmap response for the category “horse.” In contrast, the baseline method demonstrates weaker performance in handling novel categories, with responses that are weak, incomplete, or dispersed. For instance, in (b), the second column presents a relatively scattered heatmap response for the category “cat.”
Additionally, we observe that our model is capable of accurately detecting novel category objects across various complex scenes and provides precise bounding boxes for these novel categories.

\section{Conclusion}
In this paper, we propose an open-vocabulary object detection framework based on Scene-guided Relational Modeling (SRM). The framework explicitly models relational knowledge among candidate boxes within a neighbor region using a scene graph generation model and implicitly enhances category-level interactions through a global-local attention mechanism, thereby capturing the contextual associations between base and novel categories. Furthermore, in the fine-grained alignment stage, we introduce a textual alignment strategy based on image-caption pairs. A word retriever extracts category knowledge from captions and computes its similarity with relational features to guide the alignment process. Quantitative and qualitative results demonstrate that the proposed SGM framework significantly improves the detection performance of novel categories in open-vocabulary object.
\bibliographystyle{IEEEtran}
\bibliography{ref}

\newpage

\vspace{11pt}

\vspace{11pt}

\vfill

\end{document}